# Research on the Brain-inspired Cross-modal Neural Cognitive Computing Framework

**Yang Liu**

**Abstract** To address modeling problems of brain-inspired intelligence, this thesis is focused on researching in the semantic-oriented framework design for multimedia and multimodal information. The Multimedia Neural Cognitive Computing (MNCC) model was designed based on the nervous mechanism and cognitive architecture. Furthermore, the semantic-oriented hierarchical Cross-modal Neural Cognitive Computing (CNCC) framework was proposed based on MNCC model, and formal description and analysis for CNCC framework was given. It would effectively improve the performance of semantic processing for multimedia and cross-modal information, and has far-reaching significance for exploration and realization brain-inspired computing.

## 1. Introduction

The brain-inspired computing (BIC) is the integration of neurocognitive science and information technology. It would realize state-of-the-art computing system which has advanced in computing ability, efficiency and energy consumption. It constructs neural cognitive computing model, storage and processing, and explores the new generation of Brain-Inspired Intelligence (BII) in the algorithm, chip and architecture. The research contents of BIC include the brain-inspired algorithms, and brain-inspired hardware for learning and processing. For the study of brain-inspired algorithm, which one is simulated from the macroscopical overall cognitive function, the other is from the local microscopic structures of neurons, synapses and networks. But there is still a lack of effective research how to assemble advanced function of the complex system from the local network in mesoscopic.

BIC has been an obvious success at present, but it is far from reaching the general autonomous intelligence level, and its model and algorithm lack of cross-modal cognitive and multimedia process ability. There is still a long way to go to study the gap between nature intelligence and BII [1]. Until now, the study of brain-inspired model has not supported the uniform cognitive function such as perception, attention, memory, emotion, language and so on.

To address problems of cross-modal intelligence, the neural cognitive computing model based on BIC methods was researched in this paper. The study is organized as follows. Section 2 provides a research background, which deals with a literature review, including studies of BII, brain project, statistical learning, cognitive computing, neuromorphic computing and deep learning. Section 3 gives a definition of the study area, structure and mechanism of the nervous system, and function and architecture of the cognitive system. Section 4 applies the formal description for design the model for semantic computing. Finally, Section 5 concludes the chapter.

## 2. Literature Review

### 2.1. BII and brain project

The research of artificial intelligence has developed in the ups and downs of success and failure. Many countries have made the significant investment in the scientific research of artificial intelligence in recent years. In 1981, Japanese governments made the plans of the fifth generation computer with artificial intelligence, but the plan failed because of the man-machine conversation based on natural language, automatic program generation etc. in 1992. However, the research interest of scholars in artificial intelligence aroused by the success of corporate, such as DeepBlue in chess, Watson[2] in knowledge question and answer, and AlphaGo[3] [4] of DeepMind.

The methods and systems of BII are essential to achieve strong artificial intelligence. It depends on the study of brain intelligence. In 2010, the US NIH launched human connection project on macroscopic scales. The research object of human connection project is brain network. It includes two basic elements, which are the node and the connection. The node and connection are equally important, and can be defined as the neuron at microcosmic scale, the cortical column on mesoscopic scale, and the brain area on macroscopic scale. The brain activity map project was launched by the US government in 2013, which focus on functional connectivity research on mesoscopic scale. At the same time, the EU initiated the human brain project based on blue brain project. In 2014, Japan started a brain-mapping project called brain mapping by integrated neurotechnologies for disease studies. The China brain project, entitled "Brain Science and Brain-Inspired Intelligence" was launched in 2016. The basic research of China brain project is the neural circuit mechanisms of brain cognitive functions. China brain project is developing effective technology of BII, and approaches of brain disease diagnosis/intervention[5].

### 2.2. Statistical learning and cognitive computing

Bayesian theory has an indispensable role in the statistical learning. The Bayesian mechanism of the brain has been validated by a great deal of experimental result

Y. Liu (✉)
Key Laboratory of Big Data Analysis and Processing of Henan province, Intelligent Technology and Application Engineering Research Centre of Henan province, and College of Computer Science and Information Engineering, Henan University, Kaifeng 475004, P. R. China
e-mail:sea@vip.henu.edu.cn

of psychology and neurophysiology. Literature[6] presents a computational model under a Bayesian criterion that captures the human's learning abilities for a large class of simple concepts, which achieve human-level performance while outperforming recent deep learning approaches on a challenging one-shot classification task. According to Bayesian probability, causal inference and statistical theory, it can simulate the perception and cognitive process of visual and aural, which can construct a unified cognitive theoretical framework.

Probabilistic Graphical Model (PGM) based on graph theory and probability statistics is a powerful tool for modeling and statistical learning of complex stochastic systems. The each node of PGM only connection with the limited number of other nodes, and its local structure has the small world characteristic. In recent years, there is more efficient model of PGM such as topic model, and hierarchical Bayesian model based on Gaussian and Dirichlet processes. The topic model is a Bayesian generative model, which has been widely applied in text retrieval and machine vision. Many models had been introduced in the multimedia semantic processing, such as probabilistic latent semantic analysis, Latent Dirichlet Allocation (LDA), author topic model and hierarchical topic models[7]. The mostly general method is to extract the local semantic features of the media by the bag of words model, and then Bayesian reasoning was employed to intelligence analysis and statistical of the topic of the multimedia content. In addition, multimedia knowledge graph[8] based on ontology or semantic network has been used extensively in the information retrieval and content-based recommendation.

As a method, cognitive computing has existed for a long time, but it has been making a breakthrough in recent times. Literature[9] seeks nothing less than to discover, demonstrate, and deliver the core algorithms of the macaque monkey brain. As a new generation of intelligent systems, cognitive computing mainly simulated the human brain from the functional level of cognitive ability. So IBM Watson [10] based on Deep QA and transfer learning[11] to simulate cognitive processes of the human brain such as learning, thinking and decision making.

## 2.3. Neuromorphic computing and deep learning

Neuromorphic computing is modeling and simulation of bionic brain in function and structure. It mimics neuro-biological architectures by very-large-scale integration systems containing electronic analog circuits. Spiking neurons model includes Hodgkin-Huxley, integrate and fire, spike response model and Izhikevich[12] etc.. Spiking Neural Networks (SNN) [13] was employed to simulate synaptic plasticity. SNN is also known as the third generation of neural network models, which increasing the level of realism in a neural simulation. Besides neuronal and synaptic state, SNN also incorporated the concept of time into their operating model. It enabling high-precision and low power consumption, but it is still lack of impactful training algorithm.

At present, BII has achieved remarkable achievements in the neuromorphic computing. A large number of brain-inspired chips have been designed, such as TrueNorth[14] , Memristor[15], Neurogrid[16], SpiNNaker[17], Darwin [18], DianNao and DaDianNao[19]. The much brain-inspired software system has been developed, such as SpikeNET, NEURON, GENESIS and NEST. Compass of IBM company is a multi-threaded; massively parallel functional simulator and a parallel compiler that mapping a network of Long Distance Pathways (LDP) in the brain to TrueNorth. Literature [20] present neuron models of the brain as semantic pointer architecture unified network with Nengo, which can simulate the human tasks, such as image recognition, serial working memory, reinforcement learning, counting, question answering, rapid variable creation, and fluid reasoning.

Deep learning is also known as feature learning, which is a combination of big data and high-performance computing. The models that are to be trained have a lot of parameters in classic backpropagation algorithm, which increased the risk of overfitting when not enough training data exist. On the other hand, the method of random initialization of network parameters is used, and the gradient descent algorithm is poor since the gradient diffusion. In addition, it is easy failure because the algorithm traps into local optimum in non-convex optimization problems. The researchers such as Geoffrey Hinton, Yoshua Bengio, Yann LeCun and Andrew Ng believe that unsupervised learning can contribute to training of Deep Neural Network (DNN) [21]. They improve the training problems of the DNN with some ticks such as unsupervised pre-training, dropout, large-scale labeled data, rectified linear unit activation function and GPU parallel acceleration.

In the traditional model of classification and prediction, we need to train the classifier according to the features of prior knowledge or artificial extraction. The artificial features would directly affect the results, so ability of generalization and representation in classification and recognition is limited. Deep learning extracts feature and learning representations based on a set of algorithms that attempt to model high level abstractions in the data by using a deep graph with multiple layers, composed of multiple linear and non-linear transformations. There are many successful models of deep learning such as Deep Belief Network (DBN), stacked auto-encoder and Convolutional Neural Network (CNN).

Yahn Lecun proposed training neural network architecture of LeNet5 based on Neocognitron[22] for multilayer CNN structure, but the effect is not ideal. Hinton believes that simple supervised learning cannot resolve complex problems of machine learning. He proposes extraction pattern as the goal by the neural network, and development foundation of the neural network mechanism with internal representing of the external environment. In 2006, Hinton structured DBN with a double layered structure of the restricted Boltzmann machine and Bayesian belief network. In literature[23], it is believed that both the cortical structure and the cognitive process have the depth of structure, and then a fast algorithm for deep learning is offered. With the success of the 8 layers AlexNet[24], 16 layers VGGNet (OxfordNet)[25], 22 layers of GoogLeNet[26], 152 even 1001 layers of ResNet[27], deep learning has achieved excellent results in the image classification and speech recognition.

At present, deep learning and traditional machine learning algorithms are combined to solve the problem of existing computing systems which have high energy consumption and the lack of intelligence. For example, AlphaGo[28,4] improved the performance of the game program with CNN, reinforcement learning and Monte Carlo tree search algorithm. Literature[29] proposed the Network-in-Network architecture, which using 1×1 convolutions after the regular convolutional layers. It is mathematically equivalent to using a MLP after the convolutional layers. In literature[30], DeepID2 model was proposed for face recognition, and face recognition rate is 99.15% on the LFW data sets. In literature[31] proposed ResNet to speech recognition, and obtains the word error rate breakthrough on Switchboard data set, which is a low of 6.3%. In addition, some hybrid integration framework was proposed based on supervised pre-training of DBN and Hierarchical Latent Dirichlet Allocation(HLDA)[32,33]. With the development of deep learning, cognitive mechanism was employed, such as perception, attention, memory and emotion, and more and more novel neural networks were designed. For example, it can greatly improve the performance of machine translation and intelligent retrieval in natural language processing by employed recurrent neural network, long short-term memory, gated recurrent unit, and encoder-decoder model[34].

## 3. Nervous system and cognitive architecture

The target semantic in visual scene or auditory scene is the problem of complex computing. It is an essential reference to realize semantic computing of target recognition, and multimedia intelligent processing that the function of cognitive framework and the neural mechanism. With the rapid development of deep learning and cognitive computing in the field of artificial intelligence, more and more heuristic algorithms based on biological intelligence have emerged. However, there are essentially different from research object, method and implementation among neuroscience, cognitive science and computational science. In order to solve the complex problem of audio-visual semantic computation, the computational model is established urgently for simulating the brain, inspired by the framework of cognitive function and the mechanism of neural processing.

Currently, there are two main aspects which have attracted much attention in BIC. The first is to simulate cognitive function based on systematic behavioral, and the second is to research neural mechanisms based on structures of neurons, synapses, or local networks.

**Definition 1.** *Multimedia neural cognitive computing (MNCC) is to construct multimedia information processing model and algorithm for the purpose to solve the problems of semantic processing for unstructured, massive, multi-modal, spatial and temporal distribution of multimedia information. MNCC model establishes a new generation of the multimedia information processing model and algorithms with cognitive computing of system behavior in macroscopic level, and neural computing of physiological mechanisms in microscopic level.*

**Definition 2**. *Cross-modal Neural Cognitive Computing (CNCC) built brain-inspired intelligence of cross-modal information processing framework and method based on MNCC model. CNCC mainly solves the problems of cross-modal semantic computing by mechanism of multisensory integration and multimodal cooperative cognitive.*

MNCC model is a kind of BII technology for multimedia processing based on the structure and mechanism of the nervous system, and the function and framework of the cognitive system. It is an important field to realize the intelligent processing of multimedia data. Aiming at the bottleneck of cross-modal semantic computing, this paper focuses on the hierarchical CNCC framework and algorithm based on cognitive computing and deep learning.

### 3.1. Structure and mechanism of the nervous system

It is the resource of BII theory that the structure and mechanism of the nervous system, and function and the framework of the cognitive system. There is a systematic study of the brain's information processing mechanism in different disciplines. On the one hand, the neural science analysis mechanism of neural processing at the levels of the cortical structure and the neural circuits based on the white box. On the other hand, cognitive science research model of brain's information processing through cognitive function and the phenomenon based on the black box. However, computational science realizes logic computation of a finite state machine based on Turing machine model. Although computational science has made great advances, the principle and structure are essentially different between brain and Turing machines.

The structure and function of the brain are one of the most complex systems. Generally, the neocortex of the cerebellum is the core component of intelligent processing in the field of neurocognitive science; the thalamus is the switch of information entry and selective attention; the hippocampus and the limbic system are the controller for memory and emotion. The human central nervous system is composed of white matter and grey matter, and has the obvious symmetry and contralateral. The neocortex structure of the gray matter is similar to the digital analog electronic processing unit with processing linear and nonlinear function. The LDP of the white matter made up the complex White Matter Network (WMN). It can be regarded wiring diagram of neural processing. So the neocortex structure and WMN are very important for understanding the overall structure of the brain.

1) Cortical model

The nervous system model includes three types such as description model, mechanism model and interpretation model. The description model quantitatively describes research object based on the experimental data. Mechanism model simulation research objects how to run. The interpretation model explores the basic principles of the object, and the construction of the object why so run. There are common nervous system models such as neural model, synaptic model, cortical model and structural model of the nervous system and so on.

According to evolutionary hypothesis of the triune brain[35], it divides the cerebral cortex into three categories: archicortex, paleocortex and neocortex. The archicortex originates from motor brain (reptilian) layers is not clear, the paleocortex of the emotional brain (paleomammalian) lies in limbic system consists of 3 layers of neurons. The neocortex consists of 6 layers of neurons, which accounted for 90% of the area of rational brain (neomammalian). The neocortex can be divided into primary areas, secondary areas, association areas in function.

**Hypothesis 1**. *In view of the similarity neocortical structures, it can be assumed that the information processing mechanism of the neocortex is universal in the brain areas. The audio-visual and other sensory procession can be modelled by uniform cortical function, and it can be applied to prediction, learning, reasoning and other problems.*

Most studies suggest that neocortex has the similar structure in vision area, audition area and association area. Cortical columns are a basic unit for information processing in neocortex. Cortical columns have the phenomenon of hierarchical processing and the mechanism of lateral inhibition of each other. Micro-columns consist of local circuits in neocortex. Physical stimuli are perceived and encoding to generate a nerve impulse by visual-auditory sensory neurons. The micro-column is feature detection, and macro-column or super-column makes up of micro-columns to process special information and generates some cognitive functions. The spiking probability is propagating among micro-columns. Micro-columns collect information from lower neighbor micro-columns, and disseminate information from upper neighbor micro-columns. At the same time, it also receives feedback information from LDP, and prediction information from upper neighbor.

2) Long distance pathways and neural circuits

In network structure of the brain, the frontal lobe has the core nodes, and the thalamo-cortical projection system is the key connection. Both human visual system and human auditory system have a dual stream model: dorsal and ventral pathway. "what" is happening in the dorsal pathway, and "where" is happening in the ventral pathway.

The hierarchy of cortical function and structure of neural pathways and circuits can provide significant evidences for the neural cognitive model. Neural circuits are the important material of relevance such as feedback, stochastic resonance, recurrence iterative, resonance, memory, emotion, attention, language and thinking. It can provide important evidence for build neural cognitive model that the hierarchical structure of cortical function, and the structure between the neural pathways and loops. Studies indicate the network of central nervous is a scale-free and small-world complexity.

The high accuracy human brain LDP database based on the experimental data taken from references[36,37] was constructed. Fig. 1 is the WMN visualization effect of the 48 areas which consider the connection weights and nodes size. There are a large number of long distance loops in the human brain WMN, and there are a large number of connections between the thalamus and the cortex. So the relay nuclei of the thalamus are a controller of the selective attention of sensory information, and the association nuclei of the thalamus are an exchange of cortical processing information. On the basis of the study above, Fig. 2 is a simplified structure of the whole brain WMN, and it's the WMN structure model.

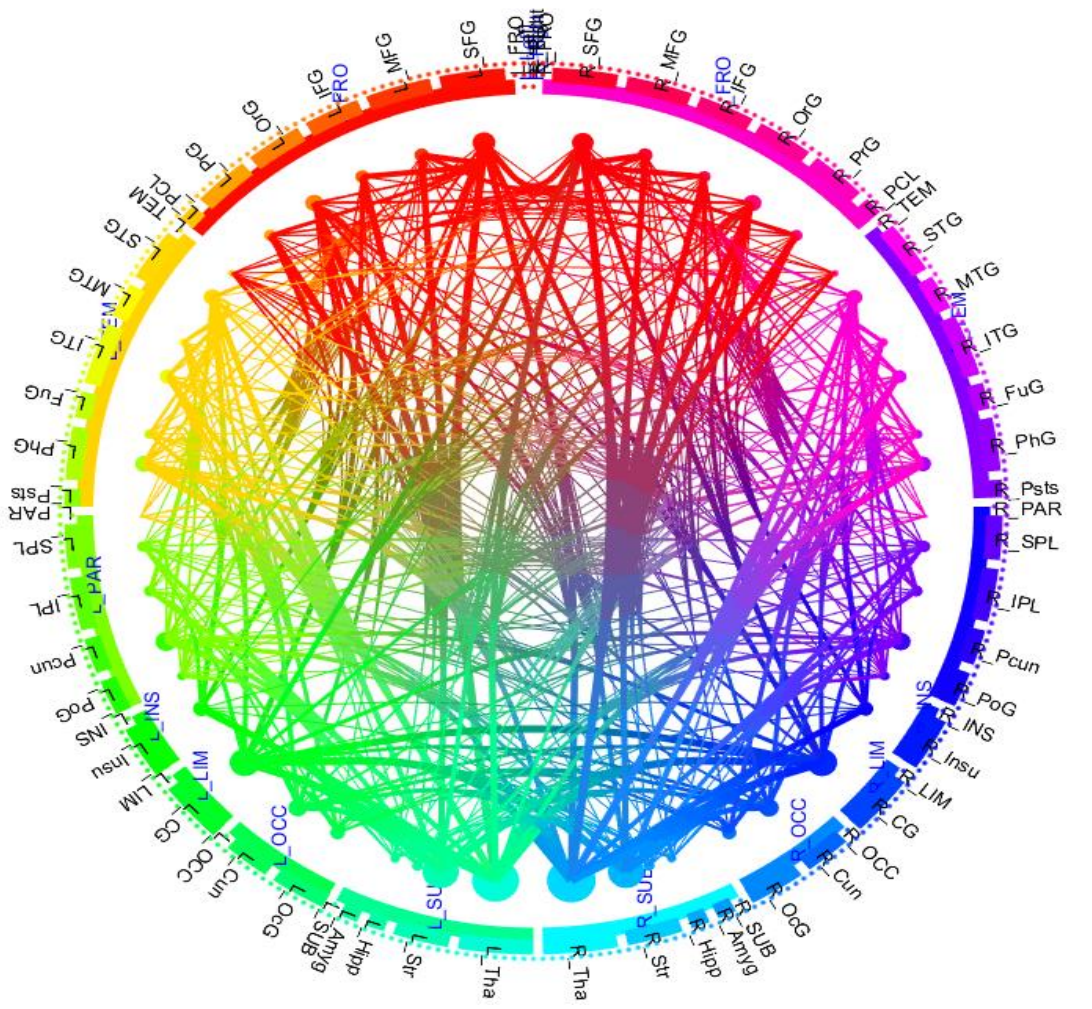

Fig. 1. WMN structures with connection weights visualization on 48 areas.

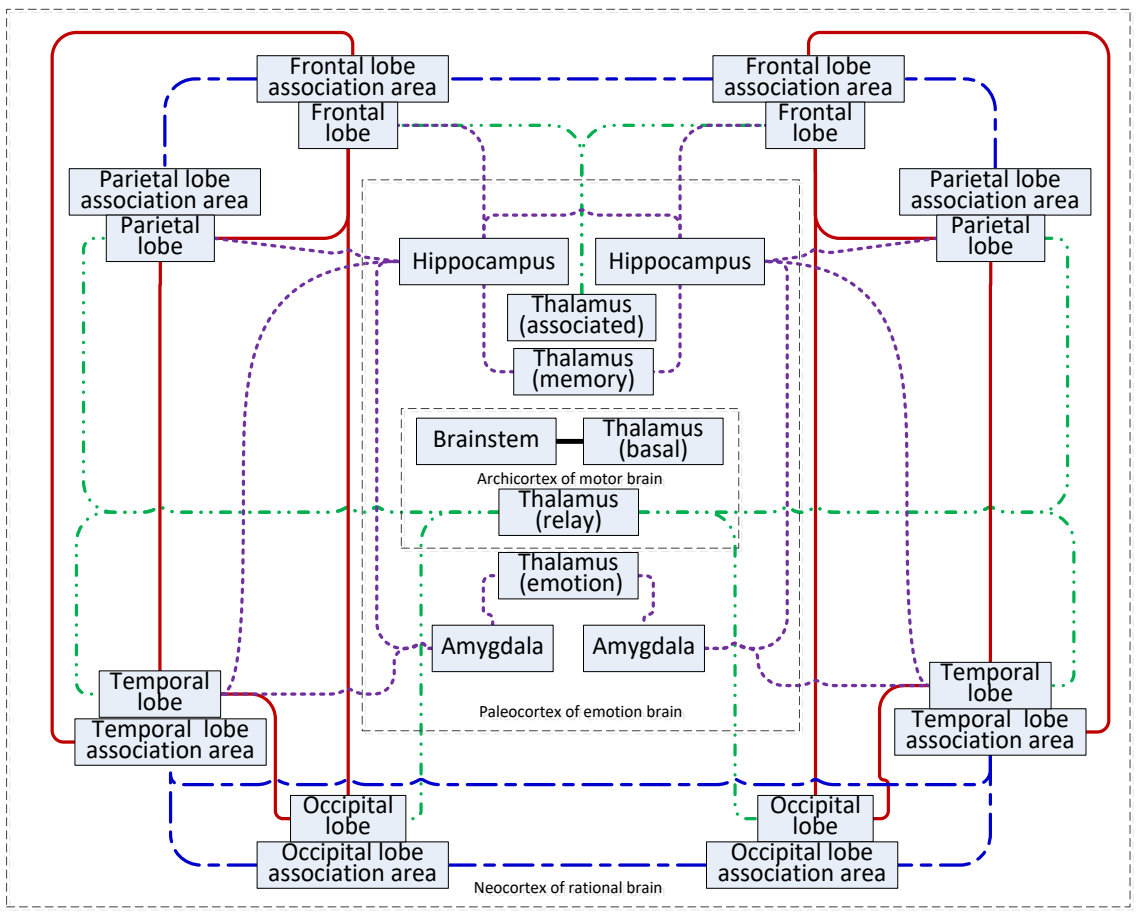

Fig. 2. WMN structure model.

## 3.2. Function and architecture of the cognitive system

Generally, the human's cognitive activities involve many aspects, for example, sense, perception, attention, emotion, memory, learning, language, thinking, representation, consciousness, knowledge representation and so on. Cognitive science explored and research on human's thinking mechanism, especially the processing mechanism by constructing cognitive model. It also provides a new architecture and technology for the design of intelligent systems.

1) Cognitive theory

In cognitive psychology, there are many cognitive models such as the ACT-R (Adaptive Control of Thought-Rational), SOAR (State, Operator And Result) [38], synesthesia model, elementary perceiver and memorizer semantic network, human associative memory, ART (Adaptive Resonance Theory), GPS (General Problem Solver), PDP (Parallel-Distributed Processing)[39] and agent et al.. Among them, cognitive theory of Bayesian probabilistic and PMJ (Perception, Memory and Judgment) model[40] should also deserve our attention.

Since the British mathematician Bayesian proposed

the probabilistic theory in 1963, the probability reasoning and decision-making of the uncertainty information had become an important content of the researches on the objective probability and cognitive processing. It is the mainstream method of machine learning and reasoning depending on the uncertainty representation of the probability, the Bayesian rule and the extension model. Cognitive researchers use Bayesian brain model[41] to simulate the cognitive process of the human brain. It is investigated cognitive processing law of subjective probability estimation by a probability model. Bayesian brain theory holds that the brain is a kind of predictive machine, and cognition is the process of probability calculation.

Cognitive researchers think that between the cognitive of the human brain and the computer information is similar in processing. They are establishing cognitive computing theory according to computers to simulate human cognitive processes. It is research and analyses the processes and principles of human cognition, cleared up the main stages and pathways of cognitive processes, and the relationship between cognitive processes and computational flow is established.

2) Cognitive architecture for media computing

Functional neuroimaging is an internal reflection of cognitive function, and it is a technical of studying the neural mechanism of brain cognition. According to literature [42-44] published neuroimaging database BrainMap (http://www.brainmap.org/ taxonomy) which is functional and structural neuroimaging with coordinate-based, Fig. 3 is the result of visualizing analysis of 48 cognitive functions in BrainMap. On the whole, the cognitive function of the human brain can be found in hierarchy obviously.

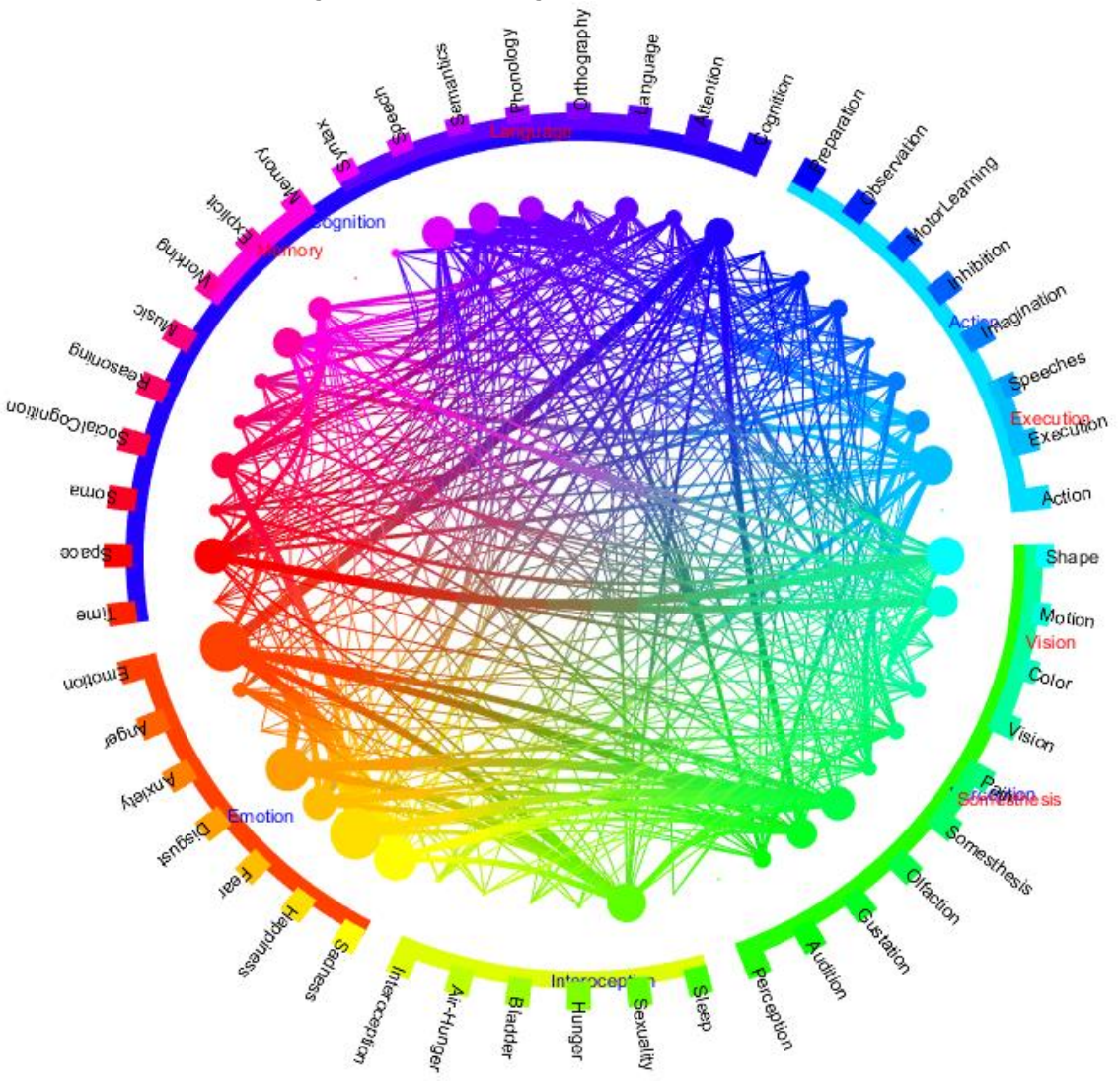

Fig. 3. Visualized analysis of 48 kinds of cognitive function.

According to the relevant function and architecture of cognition, Fig. 4 is a framework of the cognitive function for the brain. Here, the cognitive process is mainly composed of perception pathway, motion controlled pathway, attentional controlled pathway, memory and emotion circuit, feeling and decision circuit, judgment and control circuit etc.

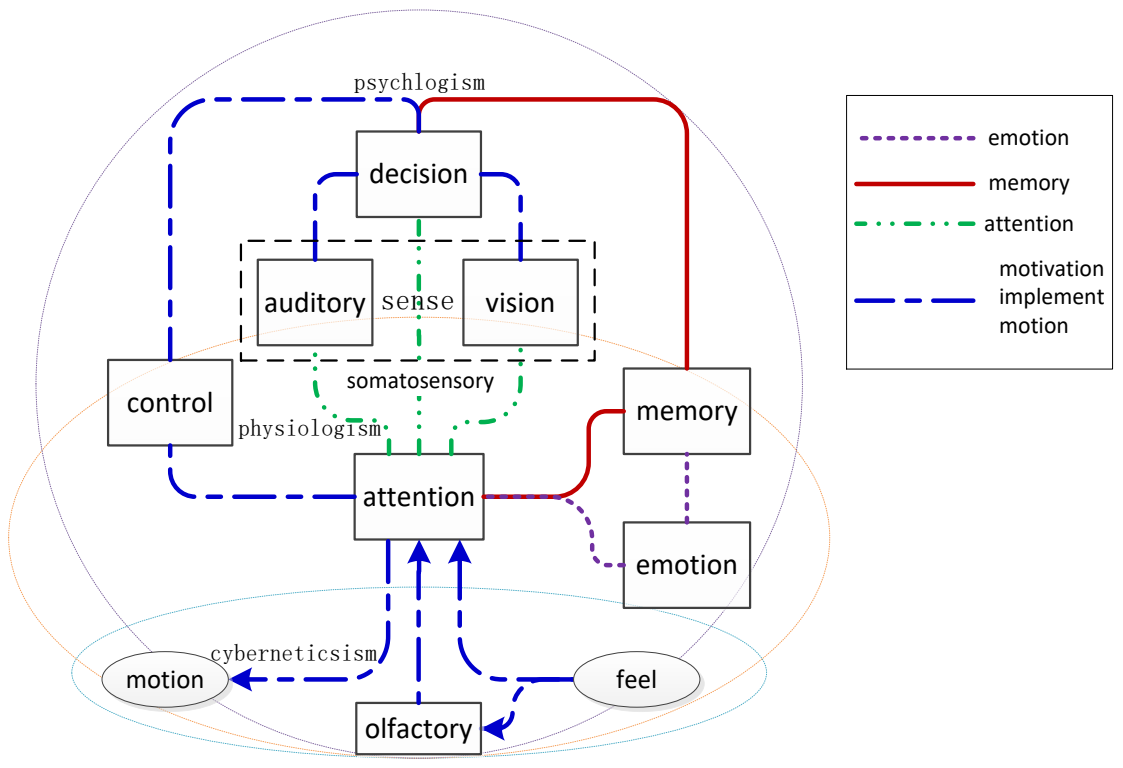

Fig. 4. Cognitive framework of brain.

## 3.3. The relationship between neural functions and cognitive functions

Fig. 5 is a visualized analysis of the correlation between the 48 neural areas and the 48 cognitive functions in the human brain LDP database according to literature[36,37].

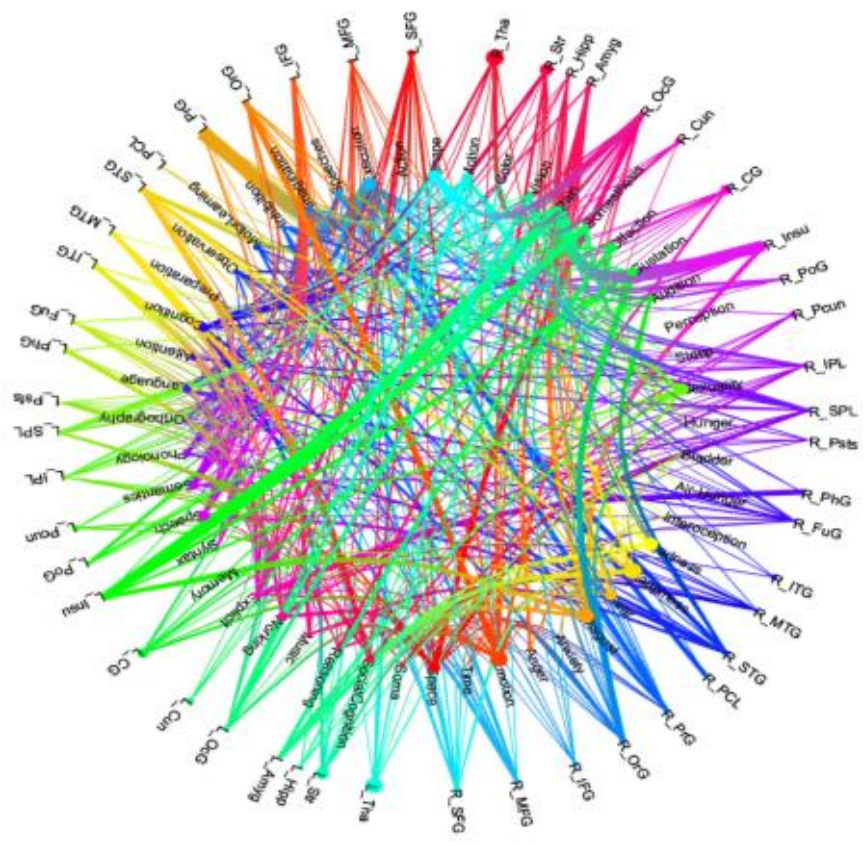

Fig. 5. Visualization and analysis between 48 neural areas and 48 kinds of cognitive function.

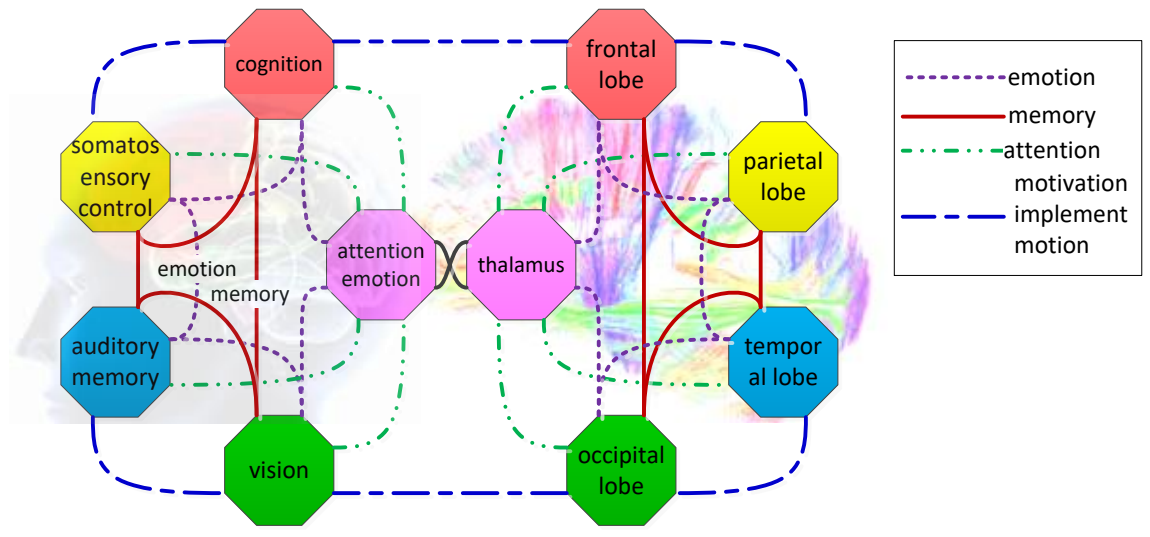

Fig. 6. Relationship between neural structure and cognitive function.

It can be noted that it is essentially fully connected between the cognitive function and neural connections. Frontal lobe and basal ganglion are the center of neural processing, and the perception and emotional are the core of cognitive function. Cognitive function is closely related to the frontal temporal lobe, the thalamus and basal ganglion are closely related to the emotion, which is the projection center of information. These results are consistent with the basic theories of neuroscience and cognitive science.

Fig. 6 shows the corresponding relationship between the neural structure and the cognitive function of the WMN in the human brain. It is an isomorphic mapping between the structure of the nervous system and function of the cognitive system in the brain. Both of them reflected the different perspectives of the brain. So hypothesis can be made as follows.

**Hypothesis 2.** *It can realize the homomorphism mapping Palatinoing between information processing (N) of nervous system and function structure (Ψ) of the cognitive system that establishment the BIC model (C) for brain (B).*

$$f: \langle \Psi, N \rangle \to C$$
$$s.t.\ N \cong \Psi \wedge C \sim B \wedge \Psi \cap N = \varphi, \Psi \subset B, N \subset B \quad (1)$$

## 4. Design of CNCC framework for semantic computing

**Hypothesis 3.** *It can realize that the object semantics recognition by BIC methods. That is, it needs to simulate the hierarchical processing, and attention mechanism of the nervous system in low-level. It also needs to simulate the function of memory and emotion in middle-level, and simulates the function of probabilistic and causality reasoning based on cognitive framework and integrated in high-level.*

This need unified hierarchical structure, for three schools such as actionism, connectionism and symbolism. The three levels for semantics-oriented computing of BII were constructed. Each of the three levels is described as follows:

**Level 1.** Perceptual computation based on control models. It simulates motor brain intelligent of perceived behavioral control on archicortex.

**Level 2.** Neural computation based on structural models. It simulates emotional brain attention circuit, emotional circuit and memory circuit of the limbic system on paleocortex. The models had incremental learning based on emotion, reinforcement learning based on memory, deep learning such as SNN, DBN and CNN etc. al.

**Level 3.** Cognitive computation based on functional models. It simulates rational brain of hierarchical ensemble learning, subjective Bayesian cognitive learning, language, and thinking control in neocortex. This method had HMM, LDA, PGM etc. al.

The semantic-oriented MNCC model research and discovers the cortex structure of the nervous system, the network structure of white matter, and cognition of the brain, such as hierarchical processing, incremental memory, emotional reinforcement, probability ensemble and so on.

### 4.1. Semantic-oriented MNCC model

As Fig. 7 shows, a semantic-oriented MNCC model based on the neural structure and cognitive framework were proposed. The model is designed based on the characteristics of neural cognitive information processing such as hierarchical processing, forward and feedback, and parallel. It extracts the semantic label from representation media by multiple steps such as region of interest extraction, saliency target detection, object-oriented incremental recognition, multi-scale target reinforcement, hierarchical ensemble process and other steps.

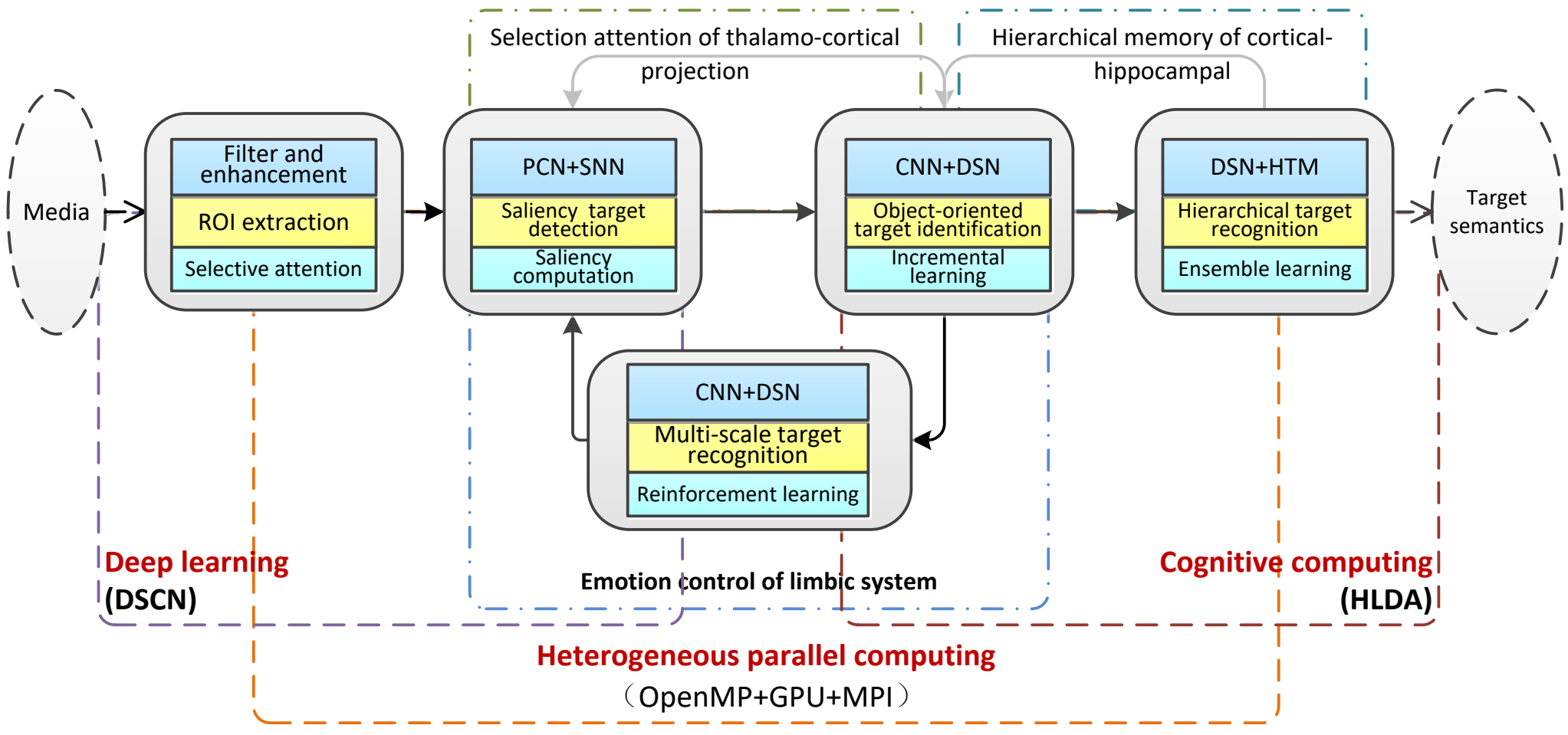

Fig. 7. Semantic-oriented MNCC model based on BIC.

In view of the hierarchical nature of the natural signals such as audio and video, high level features can be achieved through the combination of low-level features. There is the hierarchy in the language text such as word, sentence, paragraph and paper. There is the hierarchy in speech sound, for instance, sampling, phoneme, syllable and words. There is the hierarchy in the natural image, for example, pixel, edge, shape, texture, and object. The information processing of cognitive function and neural structure also had the same hierarchy from the related research of the cognitive science and neuroscience. Considering the hierarchy of neural cognitive for targets, Fig. 8 is the hierarchical CNCC framework based on MNCC for our further improvement. The hierarchical CNCC framework is designed based on the hypothesis as follows.

**Hypothesis 4**. *It can simulate low-layer perception computing process based on saliency mechanism and swarm intelligence. It can simulate the middle-layer of hierarchical feature computing process based on deep learning, reinforcement learning and incremental learning. It can simulate high-layer hierarchical decision process based on probability reasoning, causality reasoning and ensemble learning.*

### 4.2. Hierarchical CNCC framework

The goal of CNCC mainly solves the problems of multimodal semantic and cross-modal computing. There are four layers in the hierarchical CNCC framework. It includes seven sub-layers and one mixed-layer, which can realize the function of semantic computation.

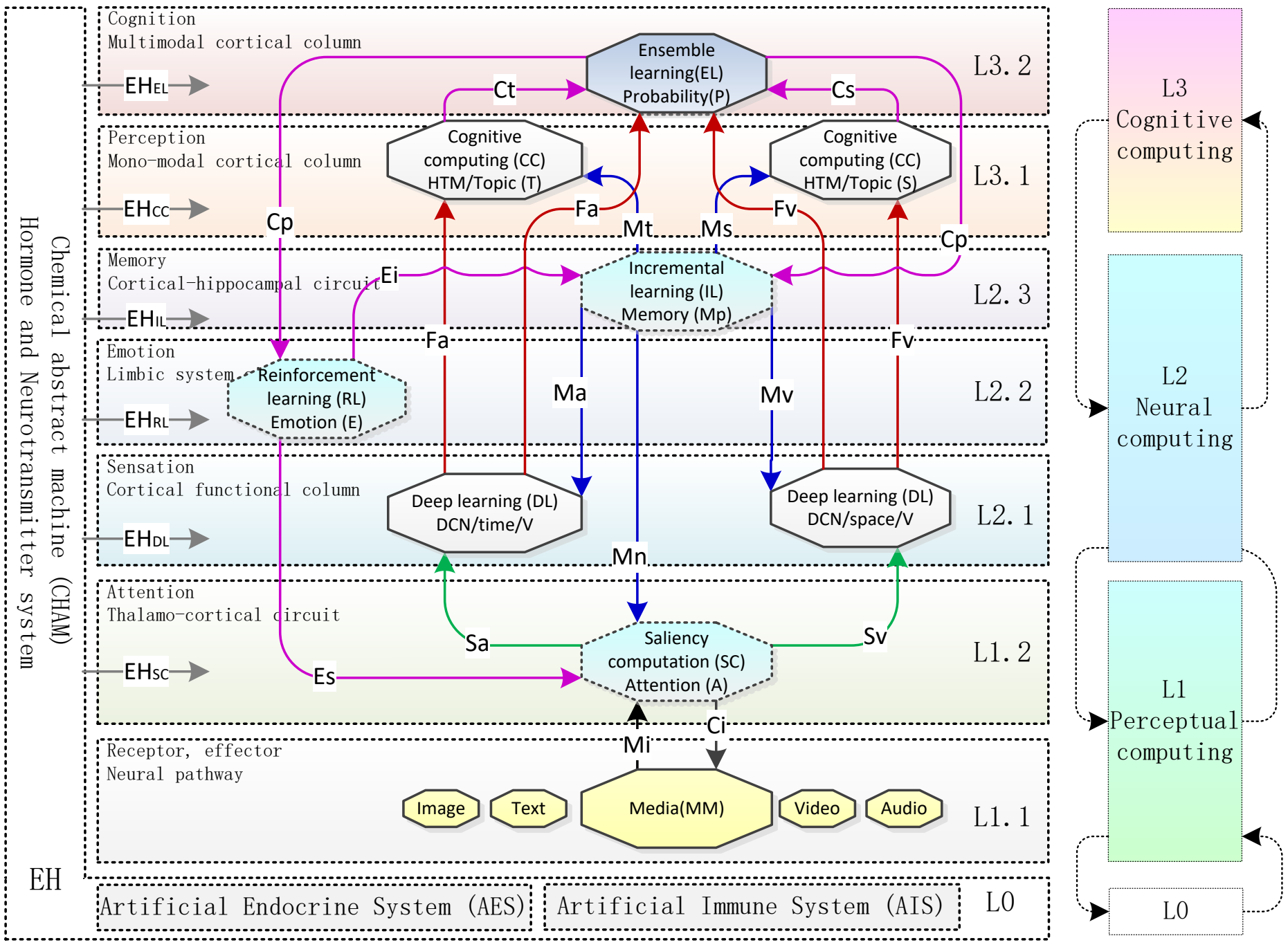

Fig. 8. The hierarchical CNCC framework based on MNCC.

**Layer 0. Mixed layer**

It is design and implements the dynamic I/O regulation according to the priori rules and inhibition and excitation mechanism of Artificial Endocrine System (AES) and Artificial Immune System (AIS).

**Layer 1. Perceptual computing layer**

**Sub-layer 1.1** It realizes pre-processing of perceptual information;

**Sub-layer 1.2** It simulates attention mechanism of the thalamus-cortical circuit, and extracts the saliency features from the media target based on sparse representation.

**Layer 2. Neural computing layer**

**Sub-layer 2.1** It simulates hierarchical structure of cortical columns, and constructs the semantic classifier based on deep learning;

**Sub-layer 2.2** It simulates the emotional reward and punishment mechanism of the limbic system, and realizes the function of the semantic reinforcement learning;

**Sub-layer 2.3** It simulates the memory mechanism of the cortex-hippocampus system, and realizes the function of the incremental semantic learning.

**Layer 3. Cognitive computing layer**

**Sub-layer 3.1** It achieves semantic cognition computing based on the theory of mono-modal cortical column and Bayes subjective probability;

**Sub-layer 3.2** It achieves semantic ensemble learning of multiple classifiers based on information integration multi-modal cortical column.

The hierarchical CNCC framework was described by following the 8-tuple <MM, SC, EL, IL, RL, DL, CC, EH>. It is mapping processing that the CNCC framework training and recognition, which can be described as follows.

$$\mathrm{MNCC}:\left\langle \mathrm{MM, SC, EL, IL, RL, DL, CC, EH} \right\rangle \rightarrow \mathrm{Cp} \quad (2)$$

The symbols and illustration of the hierarchical CNCC framework in Fig. 8 and Formula 4 are shown in Table 1.

**TABLE 1**: The symbol of CNCC framework and its implication.

| Symbol | Types | Implication |
|---|---|---|
| MM | Multidimensional matrix set | Media set, include image,audio,text,and video |
| Ma | matrix | Media |
| SC | The procedure of algorithms | Saliency computation |
| Sa | matrix | Temporal saliency feature (sparse representation) |
| Sv | matrix | Spatial saliency feature (sparse representation) |
| DL | The procedure of algorithms | Deep learning algorithm |
| IL | The procedure of algorithms | Incremental learning algorithm |
| RL | The procedure of algorithms | Reinforcement learning algorithm |
| EL | The procedure of algorithms | Ensemble learning algorithm |
| CC | The procedure of algorithms | Cognitive computing algorithm |
| Cp | set | Target semantics |
| Ct | vector | Features of temporal perception (probability topic) |
| Cs | vector | Features of spatial perception (probability topic) |
| Fa | matrix | Features of temporal senses (probability distribution) |
| Fv | matrix | Features of spatial senses (probability distribution) |
| Ma | matrix | Time increment of DNN |
| Mv | matrix | Space increment of DNN |
| Mt | vector | Time increment of cognitive topic |
| Ms | vector | Space increment of cognitive topic |
| Mp | vector | Feedback information of incremental learning |
| Mn | vector | Attention increment of saliency computation |
| Es | parameter | Reinforcement feedback of saliency computation |
| Ei | parameter | Incremental feedback of memory |
| EH | set | Endocrine molecules which effect on I/O |
| SS | set | Semantic state of chemical solution |
| TS | mapping | Reaction rule |

## 4.3. Formal description of CNCC framework

1) Formal semantics of CNCC framework based on CHAM

The mind of the brain is the physical and chemical reaction of biology. In order to analyze rationality of CNCC architecture, the Chemical Abstract Machine (CHAM)[45] was employed. CHAM describes system architecture with molecules EH (e.g., hormones, neurotransmitters and receptors), solution SS (e.g., state, semantic) and rules TS (e.g., knowledge, association, mapping). CHAM is a kind of description language architecture for parallel and dynamic software architecture analysis and testing. The CHAM molecular EH denoted factors of the chemical systems such as hormones, receptors and transmitters, which affect the function of the physical system in the nervous system.

EH=EHSC, EHDL, EHRL, EHIL, EHCC, EHEL

The connecting elements *C*, processing elements *TS* (such as knowledge, rule, association and mapping) and data elements *D* was defined as follows:

M::=TS|C◊EH|EH◊C|EH◊EH

C::=i (D)|o (D)|g (EH)|d (EH)

TS::=SC|IL|EL|RL|DL|CC

D::=Mi|Sa|Sv|Fa|Fv|Ma|Mv|Ms|Mt|Mp|Ei|Es|Cs|Ct|Cp|EH

where *I*(.) denoted the input, *O*(.) denoted the output, *G*(.) denoted the effects on the system input of the generation of hormones and transmitters, *D*(.) denoted the effects on the system output of the receptors receiving hormone and transmitter. The initial solution *SS* was defined as follows:

SS=SSSC//SSDL//SSCC//SSEL//SSRL//SSIL

where sub-solution is denoted as follows:

SSSC={|i (Mi)◊i (Mn)◊i (Es)◊g (EHSC)◊SC◊o (Sa)◊o (Sv)◊d (EHSC)|}

SSDL={|i (Sa)◊i (Ma)◊g (EHDL)◊DL◊o (Fa)//i (Sv)◊i (Mv)◊DL◊o (Fv)◊d (EHDL)|}

SSRL={|i (Cp)◊g (EHRL)◊RL◊o (Ei)◊o (Es)◊d (EHRL)|}

SSIL={|i (Ei)◊i (Cp)◊g (EHIL)◊IL◊o (Mp)◊o (Ma)◊o (Mv)◊o (Mt)◊o (Ms)◊d (EHIL)|}

SSCC={|i (Fa)◊i (Mt)◊g (EHCC)◊CC◊o (Ct)◊d (EHCC) // i (Fv) ◊i (Ms)◊g (EHCC)◊CC◊o (Cs)◊d (EHCC)|}

SSEL={|i (Ct)◊i (Cs)◊i (Fa)◊i (Fv)◊g (EHEL)◊EL◊o (Cp)◊d (EHEL)|}

The intermediate solution SM after reaction was defined as follows:

SM=SMSC//SMDL//SMCC//SMEL//SMRL//SMIL

where the sub-solution is denoted as follows:

SMSC={|SC◊i (Mi)◊i (Mn)◊i (Es)◊g (EHSC)◊o (Sa)◊o (Sv)◊d (EHSC)|}

SMDL={|DL◊i (Sa)◊i (Ma)◊g (EHDL)◊o (Fa)//DL◊i (Sv)◊i (Mv)◊o (Fv)◊d (EHDL)|}

SMRL={|RL◊i (Cp)◊g (EHRL)◊o (Ei)◊o (Es)◊d (EHRL)|}

SMIL={|IL◊i (Ei)◊i (Cp)◊g (EHIL)◊o (Mp)◊o (Ma)◊o (Mv)◊o (Mt)◊o (Ms)◊d (EHIL)|}

SMCC={|CC◊i (Fa)◊i (Mt)◊g (EHCC)◊o (Ct)◊d (EHCC) // CC◊ i (Fv)◊i (Ms)◊g (EHCC)◊o (Cs)◊d (EHCC)|}

SMEL={EL◊|i (Ct)◊i (Cs)◊i (Fa)◊i (Fv)◊g (EHEL)◊o (Cp)◊d (EHEL)|}

The six important basic rules for the solution reaction (state transition) were defined as follows:

$TS_{SC}$≡i (Mi)◊i (Mn)◊i (Es)◊g (EHSC)◊SC, o (Sa)◊o (Sv)◊ d (EHSC)◊SC→SC◊i (Mi)◊i (Mn)◊i (Es)◊g (EHSC), SC◊o (Sa)◊ o (Sv)◊d (EHSC)

$TS_{DL}$≡i (Sa)◊i (Ma)◊g (EHDL)◊DL, o (Fa)◊d (EHDL)◊DL, i (Sv)◊i (Mv)◊g (EHDL)◊DL, o (Fv)◊d (EHDL)◊DL→DL◊i (Sa)◊ i (Ma)◊g (EHDL), DL◊o (Fa)◊d (EHDL), DL◊i (Sv)◊i (Mv)◊ g (EHDL), DL◊o (Fv)◊d (EHDL)

$TS_{RL}$≡i (Cp)◊g (EHRL)◊RL, o (Ei)◊o (Es)◊d (EHRL)◊RL→RL◊i (Cp)◊g (EHRL), RL◊o (Ei)◊o (Es)◊d (EHRL)

$TS_{IL}$≡i (Ei)◊i (Cp)◊g (EHIL)◊IL, o (Mp)◊o (Ma)◊o (Mv)◊ o (Mt)◊o (Ms)◊d (EHIL)◊IL→IL◊i (Ei)◊i (Cp)◊g (EHIL), IL◊ o (Mp)◊o (Ma)◊o (Mv)◊o (Mt)◊o (Ms)◊d (EHIL)

$TS_{CC}$≡i (Fa)◊i (Mt)◊g (EHCC)◊CC, o (Ct)◊d (EHCC)◊CC, i(Fv)◊i (Ms)◊g (EHCC)◊CC, o (Cs)◊d (EHCC)◊CC→CC◊i (Fa)◊ i (Mt)◊g (EHCC), CC◊o (Ct)◊d (EHCC), CC◊i (Fv)◊i (Ms)◊ g (EHCC), CC◊o (Cs)◊d (EHCC)

$TS_{EL}$≡i (Ct)◊i (Cs)◊i (Fa)◊i (Fv)◊g (EHEL)◊EL, o (Cp)◊ d (EHEL)◊EL→EL◊i (Ct)◊i (Cs)◊i (Fa)◊i (Fv)◊g (EHEL), EL◊ o (Cp)◊d (EHEL)

The rules $TS_{SC}$ denoted the saliency computation of the attention mechanism in thalamic-cortical circuits. Thalamic-cortical projection is an important infrastructure of brain function, and the thalamus plays an important role in the attention mechanism. Selective attention can reduce the influence of curse of dimensionality by saliency mechanism. In order to realize the saliency computation, this process focuses on the attention mechanism of the thalamic-cortical circuit, and establishes the scheme of the saliency feature extraction. This rule mapping between the media *MM* and the spatial-temporal saliency features <*Sa*, *Sv*> was indicated as follows:

$$SC : \langle MM, Mn, Es \rangle \rightarrow \langle Sa, Sv \rangle \quad (3)$$

The rules $TS_{DL}$ denotes the senses feature learning based on the hierarchical structure of the cortex. Cortical column is the basic unit of cognitive function. The cortex cognitive function is deep learning algorithm research basis, and inspires how to realize the classification and recognition of the target. We can simulate the processing mechanism of multi-layers architecture of the cortical column, and design hierarchical semantic classifier. Probability distribution of the spatial-temporal senses features<*Fa*,*Fv*> was computed with media objects saliency features as follows:

$$DL : \{\langle Sa, Ma \rangle \langle Sv, Mv \rangle\} \rightarrow \langle Fa; Fv \rangle \quad (4)$$

The rules $TS_{CC}$ denoted perceptual features computation based on probabilistic cognition. The computation process of perceptual feature is building the mapping between Bayesian probability distribution of spatial-temporal senses features <*Fa*,*Fv*> and spatial-temporal perceptual features <*Ct,Cs*>.

$$CC:\left\{\langle Fa, Mt\rangle;\langle Fv, Ms\rangle\right\} \rightarrow \{Ct;Cs\} \tag{5}$$

The rules $TS_{EL}$ denoted target recognition based on multi-modal perception integration. It realizes the Ensemble Learning (EL) of multi-modal perception information, and the final semantic decision making of target recognition. The core mission of target recognition is to establish the mapping between spatial-temporal senses-perceptual features <*Ct, Cs, Fa, Cp, Fv*> and target semantic labels as follows.

$$EL:\langle Cs, Ct, Fv, Fa \mid DL, CC\rangle \rightarrow Cp \tag{6}$$

The rules $TS_{RL}$ denoted the reward and punishment of emotion in the limbic system. It is the Reinforcement Learning (RL) basis that the emotions control of reward and punishment in the limbic system. The aim of simulating emotion control of rewards and punishment is to establish a stable and optimal target semantic. This rule solves errors minimization paradigm between the target semantic expectation *Cp* and the saliency feedback (*Ei* and *Es*) was defined as follows.

$$\arg \underset{RL}{\overset{\|RL\|}{Min}}\left(\left\|Cp_L - \overline{Cp}\right\|\right)$$
$$s.t.\quad RL:\langle Cp \mid CC\rangle \rightarrow \langle Ei, Es\rangle \tag{7}$$

The rules $TS_{IL}$ denoted the control of the memory system. The essence of semantic mapping is the memory and prediction for spatial-temporal pattern. The material base for intelligent prediction includes the memory processing architecture of cortex-hippocampus circuits, and its spatial-temporal pattern. This rule simulated mechanism of memory control, and storage and prediction of the historical information. The rules employed Incremental Learning (IL) method to control incremental information. It includes the incremental of DNN's spatial-temporal features <*Mt,Ms*>, the incremental of cognitive topic spatial-temporal features <*Mt,Ms*>, and memory feedback *Mp* of incremental learning.

$$IL:\langle Ei, Cp \mid DL, CC\rangle \rightarrow \langle Mp, Ma, Mv, Mt, Ms, Mn\rangle \tag{8}$$

2) Description and analysis for semantic learning and recognition of hierarchical CNCC framework

The dynamic process of hierarchical CNCC framework learning includes the following three steps and six basic processes as follows:

| | | |
|---|---|---|
| (1) | | It achieves spatial-temporal saliency features computation based on SNN according to the rules of $TS_{SC}$.<br>SSSC//SSDL//SSCC//SSEL//SSRL//SSIL→SMSC//SSDL//SSCC//SSEL//SSRL//SSIL |
| (2) | | It achieves target semantic learning of hierarchical integrated cognition based on deep learning and cognitive computing, including three dynamic processes as follows: |
| | (2.1) | It realizes spatial-temporal senses features computation of DNN according to the rules of $TS_{DL}$.<br>SMSC//SSDL//SSCC//SSEL//SSRL//SSIL→SMSC//SMDL//SSCC//SSEL//SSRL//SSIL |
| | (2.2) | It realizes spatial-temporal perception features computation of hierarchical topic model according to the rules of $TS_{CC}$.<br>SMSC//SMDL//SSCC//SSEL//SSRL//SSIL→SMSC//SMDL//SMCC//SSEL//SSRL//SSIL |
| | (2.3) | It realizes ensemble learning of objects semantic labels based on AdaBoost according to the rules of $TS_{EL}$.<br>SMSC//SMDL//SMCC//SSEL//SSRL//SSIL→SMSC//SMDL//SMCC//SMEL//SSRL//SSIL |
| (3) | | It achieves incremental computation and feedback of reinforcement learning based on object-oriented and multi-scale, including two dynamic processes as follows: |
| | (3.1) | It realizes multi-scale feedback computation of hierarchy reinforcement learning according to the rules of $TS_{RL}$.<br>SMSC//SMDL//SMCC//SMEL//SSRL//SSIL→SMSC//SMDL//SMCC//SMEL//SMRL//SSIL |
| | (3.2) | It realizes the temporal-spatial computation of object-oriented target based on online incremental learning according to the rules of $TS_{IL}$.<br>SMSC//SMDL//SMCC//SMEL//SMRL//SSIL→SMSC//SMDL//SMCC//SMEL//SMRL//SMIL |

The dynamic process of semantic recognition based on hierarchical CNCC framework learning includes the following two steps and four basic processes as follows:

| | | |
|---|---|---|
| (1) | | It achieves the saliency feature computation of sparse representation of SNN.<br>SSSC//SSDL//SSCC//SSEL//SMRL//SMIL→SMSC//SSDL//SSCC//SSEL//SMRL//SMIL |
| (2) | | It achieves target recognition of hierarchical integrated cognition based on deep learning and cognitive computing, including three processes as follows: |
| | (2.1) | It realizes spatial-temporal senses feature computation of DNN according to the rules of $TS_{DL}$.<br>SMSC//SSDL//SSCC//SSEL//SMRL//SMIL→SMSC//SMDL//SSCC//SSEL//SMRL//SMIL |
| | (2.2) | It realizes spatial-temporal perception features computation of hierarchical topic model according to the rules of $TS_{CC}$.<br>SMSC//SMDL//SSCC//SSEL//SMRL//SMIL→SMSC//SMDL//SMCC//SSEL//SMRL//SMIL |
| | (2.3) | It realizes ensemble computation of objects semantic labels based on AdaBoost according to the rules of $TS_{EL}$<br>SMSC//SMDL//SMCC//SSEL//SMRL//SMIL→SMSC//SMDL//SMCC//SMEL//SMRL//SMIL |

## 4.4. Applications and experimental of CNCC framework

A wide range of applications of semantic-oriented MNCC model and hierarchical CNCC framework would be identified such as unmanned autonomous systems and search engines of cross-media intelligent[46]. It would have profound significance for the exploration and the realization of the BIC. It had been applied to the algorithms of scene classification, target detection and target recognition of high resolution remote sensing images [47]. The experimental result of scene classification, target detection and target recognition based on CNCC framework and MNCC model is shown in Table 2.

The Average Precision (AP) of scene classification algorithm based on MNCC model is up to 84.73% on High-resolution Satellite Scene (HRSS) data set of Wuhan University, and reaches 88.26% on University of California Merced Land Use (UCMLU) data set.

For target detection algorithm based on MNCC model on High-resolution Remote Sensing Harbor Target Detection (HRSHTD) data set, the average Probability of Detection (PD) is 91.63%, False Alarm Rate (FAR) is 8.37%, and Missed Detection Rate (MDR) is 9.35%.

The experimental results show that AP and Overall Accuracy (OA) of target classification algorithm based on CNCC framework are 96.93% and 97.00% on data set of high-resolution target classification and recognition, respectively [47]. It also reaches 99.90% and 99.88% on Moving and Stationary Target Acquisition and Recognition (MSTAR) data set, respectively.

**Table 2**: The experimental results sematic recogntive based on CNCC framework and MNCC model.

| Task | Dataset | Medthod | Evaluation(%) |
|---|---|---|---|
| Scene classification | HRSS | MNCC | AP=84.73 |
| Scene classification | UCMLU | MNCC | AP=88.26 |
| Target detection | HRSHTD | MNCC | PD=91.63<br>FAR=8.37<br>MDR=9.35 |
| Target recognition | MSTAR | CNCC | AP=99.90<br>OA=99.88 |
| Target recognition | HSTCR | CNCC | AP=97.00<br>OA=96.93 |

It shows that the CNCC framework can address the problem of semantic learning on remote sensing images data set, which are small and complex ground objects.

## 5. Conclusion

Aiming at scientific problems of BIC modeling, the cortical models and WMN characteristic in the nervous system had been analyzed in this paper. Then hierarchy and other characteristic of architecture and function in the cognitive system had been explored. The relationship between nervous system and cognitive framework for BII had also summarized. The hierarchical CNCC framework is proposed based on the MNCC model. The formal analysis of the rationality is given for hierarchical CNCC framework by CHAM. Looking to the future, it can be applied to the cross-modal intelligence applications, to realize the state-of-the-art applications based on semantic CNCC framework are the next step.

**Acknowledgements** This work is supported by National Natural Science Foundation of China (61305042), Projects of Center for Remote Sensing Mission Study of China National Space Administration (2012A03A0939), Key Research and Promotion Projects of Henan Province (182102310724), and Science and Technological Research of Key Projects of Henan Province (172102110006).

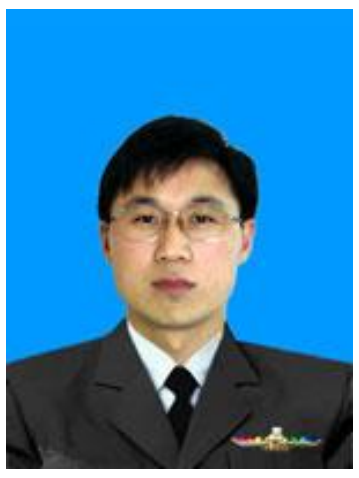

**Yang Liu** is associate professor and M.S. supervisor of college of computer science and information engineering, Henan University, China. He received M.S. and Ph.D. degrees from Henan University in 2008 and 2016, respectively. He received his B.S. degrees from Changchun University of science and technology (now it merged into faculty of science, Jilin University) in 1996. His research interests include multimedia neural cognitive computing in brain inspired computing, temporal and spatial information high-performance computing in remote sensing, and intelligent processing information in multimedia.